\documentclass[10pt,twocolumn,letterpaper]{article}

\usepackage[pagenumbers]{wacv} 

\usepackage{graphicx}
\usepackage{amsmath}
\usepackage{amssymb}
\usepackage{booktabs}

\usepackage[pagebackref,breaklinks,colorlinks]{hyperref}

\usepackage[capitalize]{cleveref}
\crefname{section}{Sec.}{Secs.}
\Crefname{section}{Section}{Sections}
\Crefname{table}{Table}{Tables}
\crefname{table}{Tab.}{Tabs.}

\def\wacvPaperID{264} 
\def\confName{WACV}
\def\confYear{2024}

\begin{document}

\title{DreamCatcher: Revealing the Language of the Brain with fMRI using GPT Embedding}

\author{Subhrasankar Chatterjee  \\
Indian Institute of Technology, Kharagpur\\
Kharagpur-721302, West Bengal, India.\\
{\tt\small subhrasankarphd@iitkgp.ac.in}
\and
Debasis Samanta\\
Indian Institute of Technology, Kharagpur\\
Kharagpur-721302, West Bengal, India.\\
{\tt\small dsamanta@iitkgp.ac.in}
}
\maketitle

\begin{abstract}
   The human brain possesses remarkable abilities in visual processing, including image recognition and scene summarization. Efforts have been made to understand the cognitive capacities of the visual brain, but a comprehensive understanding of the underlying mechanisms still needs to be discovered. Advancements in brain decoding techniques have led to sophisticated approaches like fMRI-to-Image reconstruction, which has implications for cognitive neuroscience and medical imaging. However, challenges persist in fMRI-to-image reconstruction, such as incorporating global context and contextual information. In this article, we propose fMRI captioning, where captions are generated based on fMRI data to gain insight into the neural correlates of visual perception. This research presents DreamCatcher, a novel framework for fMRI captioning. DreamCatcher consists of the Representation Space Encoder (RSE) and the RevEmbedding Decoder, which transform fMRI vectors into a latent space and generate captions, respectively. We evaluated the framework through visualization, dataset training, and testing on subjects, demonstrating strong performance. fMRI-based captioning has diverse applications, including understanding neural mechanisms, Human-Computer Interaction, and enhancing learning and training processes.
\end{abstract}


\section{Introduction}
The human brain manifests excellent proficiencies in visual processing, encompassing image recognition and scene summarization. Considerable endeavors have been devoted to elucidating the cognitive capacities inherent in the visual brain. Nevertheless, a comprehensive understanding of the fundamental mechanisms that underlie human visual processing remains an unresolved endeavor. 

\begin{figure}
  \centering
  \includegraphics[width=\linewidth]{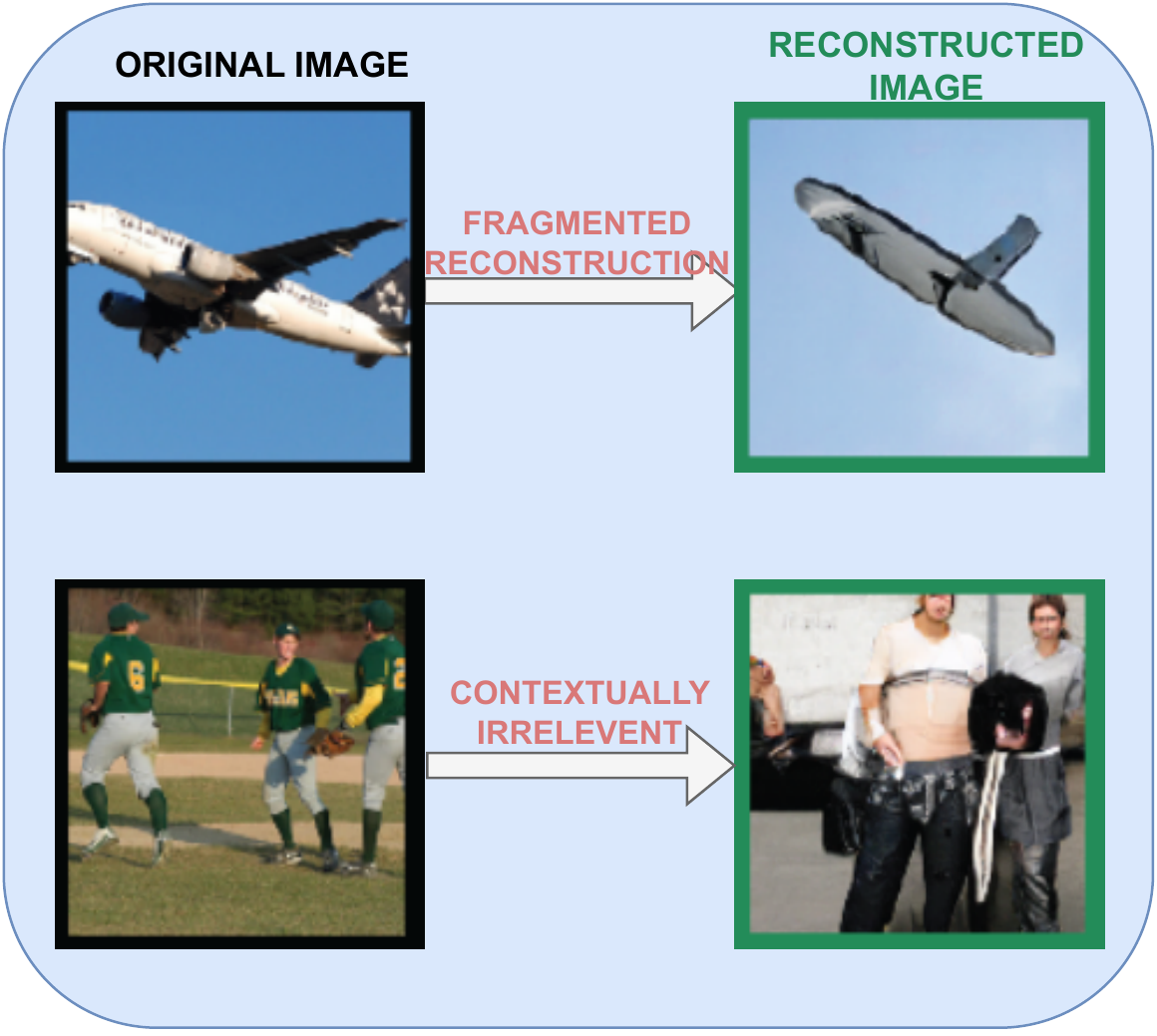}
  \caption{\textbf{Illustrative example of current issues with fMRI-to-Image reconstruction.} First reconstruction example successfully captures the low-level features however misses the high-level features. Second reconstruction example is adequate at a object-level replication, however misses the context in which the objects are to be placed.}
  \label{fig:why}
\end{figure}

\begin{figure*}
  \centering
  \includegraphics[width=0.8\textwidth]{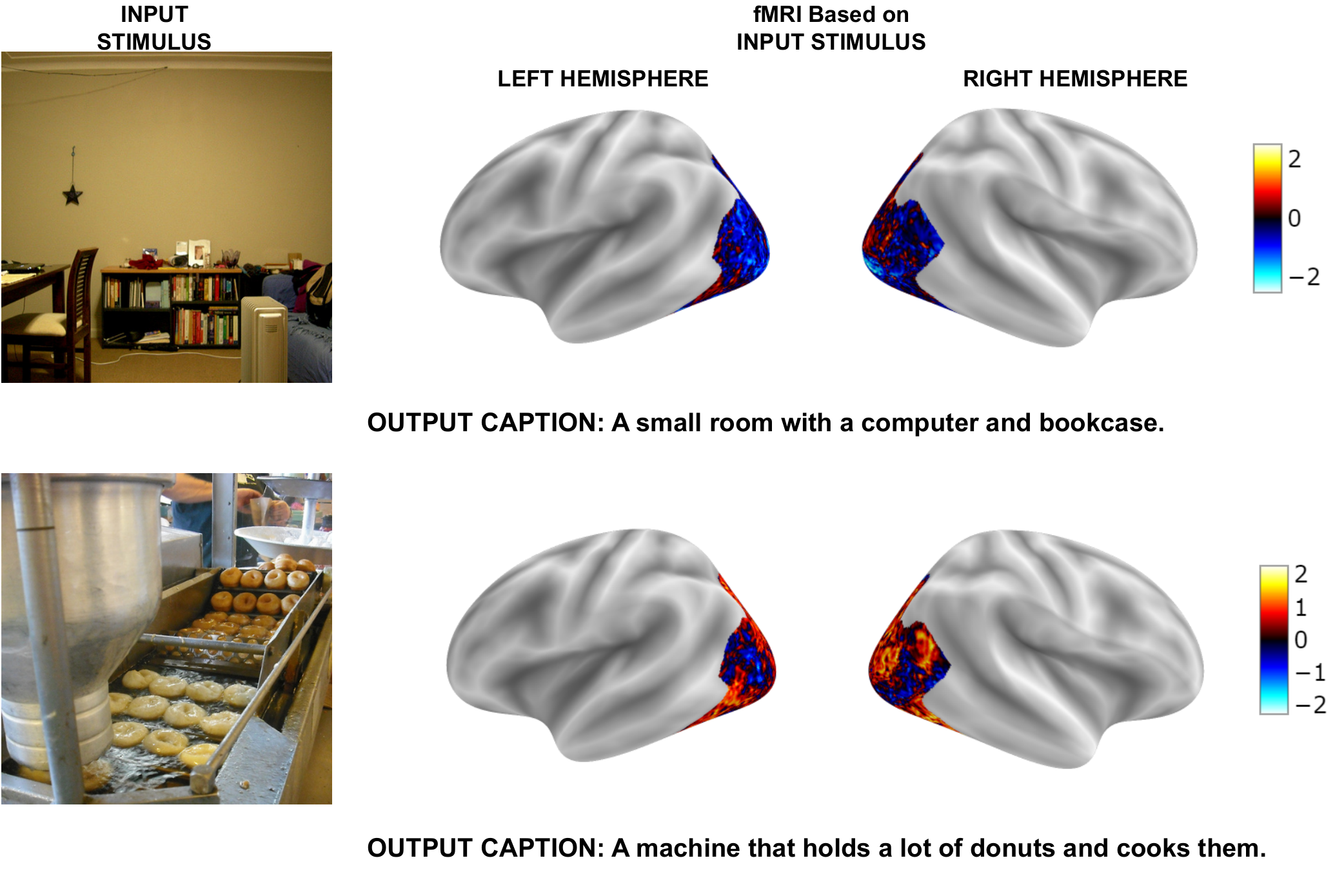}
  \caption{\textbf{fMRI-Captioning:} Subject is presented with an image stimulus and fMRI Neural Responses were captured. Given an fMRI response, the task is to predict captions based on the visual stimulus.}
  \label{fig:what}
\end{figure*}

Functional magnetic resonance imaging (fMRI) has emerged as an invaluable instrument for scrutinizing the neural activity of the human brain \cite{Haynes2006, Kamitani2005, Norman2006, Naselaris2011}. Over time, fMRI-based brain decoding techniques have progressed from rudimentary fMRI classification approaches \cite{Cox2003, Haxby2001, Kay2008, Haxby2011} to the more sophisticated realm of fMRI-to-Image reconstruction \cite{Miyawaki2009, Naselaris2009, Nishimoto2011}. This evolutionary trajectory holds significant implications for both the comprehension of neural mechanisms and the practical application of such knowledge. Particularly, domains such as fMRI-to-Image reconstruction have the potential to revolutionize fields, including cognitive neuroscience \cite{Kar2022, Kriegeskorte2018, Palacio2018} and even medical imaging \cite{Bore2022, chua2022, Randeniya2022}. These advancements paved the way for proposing a more sophisticated approach to brain decoding known as fMRI captioning(Please refer to \cref{fig:what}). In this domain, captions are generated for input stimuli based on fMRI data, affording a deeper understanding of the neural correlates of visual perception.

Deep generative models, including Variational Autoencoders (VAEs), Generative Adversarial Networks (GANs), and Latent Diffusion Models (LDMs), have significantly advanced visual reconstruction. These models have been widely used to reconstruct complete images by mapping brain signals to latent variables. Successful applications include face reconstruction \cite{VanRullen2019, Dado2022} , single-object-centered image reconstruction \cite{Shen2019} , and complex scene reconstruction \cite{Allen2022,Lin2022}. Previous studies have focused on datasets like Generic Object Decoding and Deep Image Reconstruction derived from ImageNet, and demonstrated improvements in reconstruction quality using approaches such as deep generator networks \cite{Shen2019}, supervised and unsupervised training \cite{Beliy2019}, BigBiGAN-based models \cite{Mozafari2020}, and dual VAE-GAN models \cite{Ren2021}. The Natural Scenes Dataset (NSD), curated by Allen et al. (2022) \cite{Allen2022}, has emerged as a benchmark for fMRI-based natural scene reconstruction, with studies employing models such as StyleGAN2 \cite{Lin2022}, Stable Diffusion \cite{Takagi2022}, and improved IC-GAN frameworks \cite{Gu2022} to reconstruct images and estimate pose.

Despite the advancement in architectures for fMRI-to-image reconstruction, specific inherent challenges persist. Firstly, the reconstruction process is typically fragment-based, necessitating understanding the image's global context within most frameworks. Secondly, while fragment-based reconstructions can effectively capture low-level object features, they often need help to incorporate the contextual information of the image. As illustrated in \cref{fig:why}, taken from the work of Gu et al. \cite{Gu2022}, the first part of the image demonstrates the reconstruction of an airplane, revealing the successful capture of low-level features but a failure to reproduce high-level features resulting in fragmented reconstruction. Similarly, the human objects are adequately replicated in the second part of the image, but the contextual aspects remain absent. These substantial challenges are addressed through the application of fMRI-captioning techniques.

\begin{figure*}[ht!]
  \centering
  \includegraphics[width=0.8\textwidth]{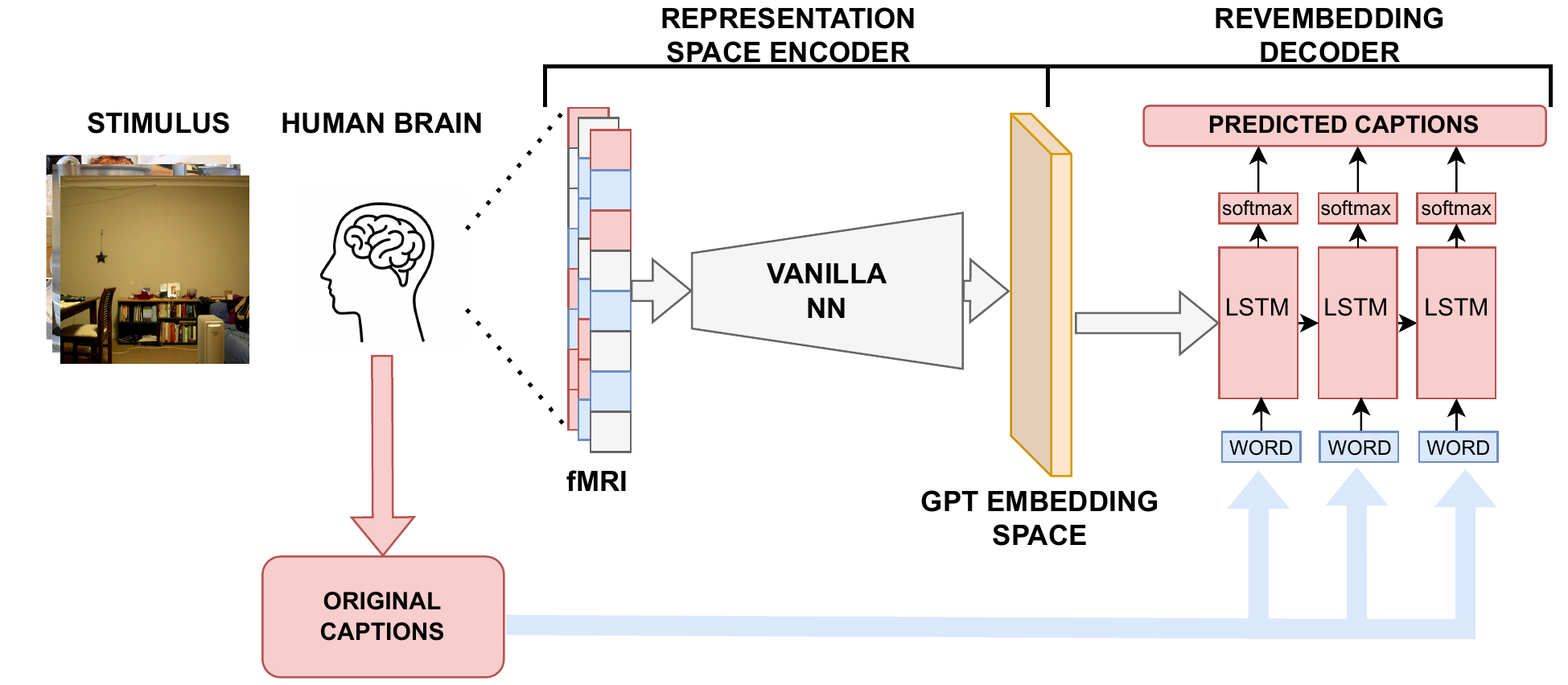}
  \caption{\textbf{Algorithmic Pipeline:} The human subject was presented with visual stimuli while fMRI was being recorded. For each visual stimuli, captions were also produced by the subject. The DreamCatcher Framework has two components: the RSE takes in the fMRI neural response and generates a 1536-D GPT Embedding, the RevEmbedding Decoder uses a pone-to-many sequential model that takes the GPT Embedding and produces a caption. The loss is calculated using the original human caption and predicted fMRI caption.}
  \label{fig:How}
\end{figure*}

In this research paper, we introduce a novel framework called DreamCatcher for fMRI captioning. The DreamCatcher framework comprises two key components: the Representation Space Encoder (RSE) and the RevEmbedding Decoder(Please refer to \cref{fig:How}). The RSE is designed as a standard neural network architecture, which takes preprocessed fMRI vectors as input and projects them onto an N-Dimensional Representation space. Our framework adopts a 1536-D GPT Embedding \cite{Neelakantan2022} as the representation space, as it is well-suited for language-based transformations. The RSE acts as a transformation function \cite{Vaswani2017} that maps the fMRI input space to a latent space based on a pre-trained Large Language Model (LLM). However, existing reverse embedding techniques often rely on approximations unsuitable for our purpose. Hence, the RevEmbedding Decoder, the second part of our model, is implemented as a One-to-many LSTM Decoder. It takes the GPT Embedding as input and generates the desired captions. To evaluate the performance of our framework, we conducted three sets of experiments. Firstly, we visualized the generated representation space to assess its ability to capture adequate representations. Secondly, we trained and tested the entire framework using the Natural Scene Dataset \cite{Allen2022} and MS-COCO Dataset \cite{Lin2014} to establish its feasibility. Since fMRI captioning is still a nascent field, we could not compare our framework directly with state-of-the-art models. Finally, we evaluated the effectiveness of our framework by testing it on two subjects, employing metrics such as METEOR, Sentence, and Perplixity. The results indicate that our DreamCatcher framework exhibits strong performance as an fMRI captioning model.

fMRI-based captioning finds application in understanding neural mechanisms and domains like Human-Computer Interaction. Researchers can design more intuitive and responsive interfaces that adapt to users' cognitive states by discerning the brain's response to diverse visual inputs. The utilization of fMRI-based caption generation has the potential to support learning and training processes, particularly in educational settings. Researchers can discern patterns associated with successful learning or engagement by analyzing neural responses to visual stimuli during educational tasks. This amalgamation of neuroimaging techniques and educational contexts holds promise for enhancing pedagogical practices through a nuanced understanding of the brain's cognitive responses to visual stimuli.

DreamCatcher effectively addresses the limitations inherent in fragment-based reconstructions by incorporating contextual information through an LSTM module. This approach enables more comprehensive and coherent reconstructions of visual stimuli, capturing not only low-level object features but also high-level contextual aspects. Notably, the DreamCatcher framework has the potential for versatility by being independent of neural response modality, making it applicable to other modalities such as EEG(Electroencephalogram) or ECoG(Electrocorticogram). This adaptability extends the potential of DreamCatcher in real-time applications.

To summarize, the contributions of this research are as follows:
\begin{itemize}
    \item \textbf{Introduction of fMRI Captioning:} This study introduces the fMRI captioning domain as an alternative approach to traditional fMRI-to-Image Reconstruction for Neural Decoding.
    \item \textbf{Proposal of the DreamCatcher Framework:} The DreamCatcher framework is proposed as a feasibility test for fMRI captioning.
    \item  \textbf{Verification of GPT Embedding Space Applicability:} This research validates the use of GPT Embedding space as a brain representation space within the DreamCatcher framework. The effectiveness of this representation space is demonstrated through empirical evaluation, providing evidence for its utility in fMRI captioning tasks.
\end{itemize}

\begin{figure*}[ht]
  \centering
  \includegraphics[width=0.8\textwidth]{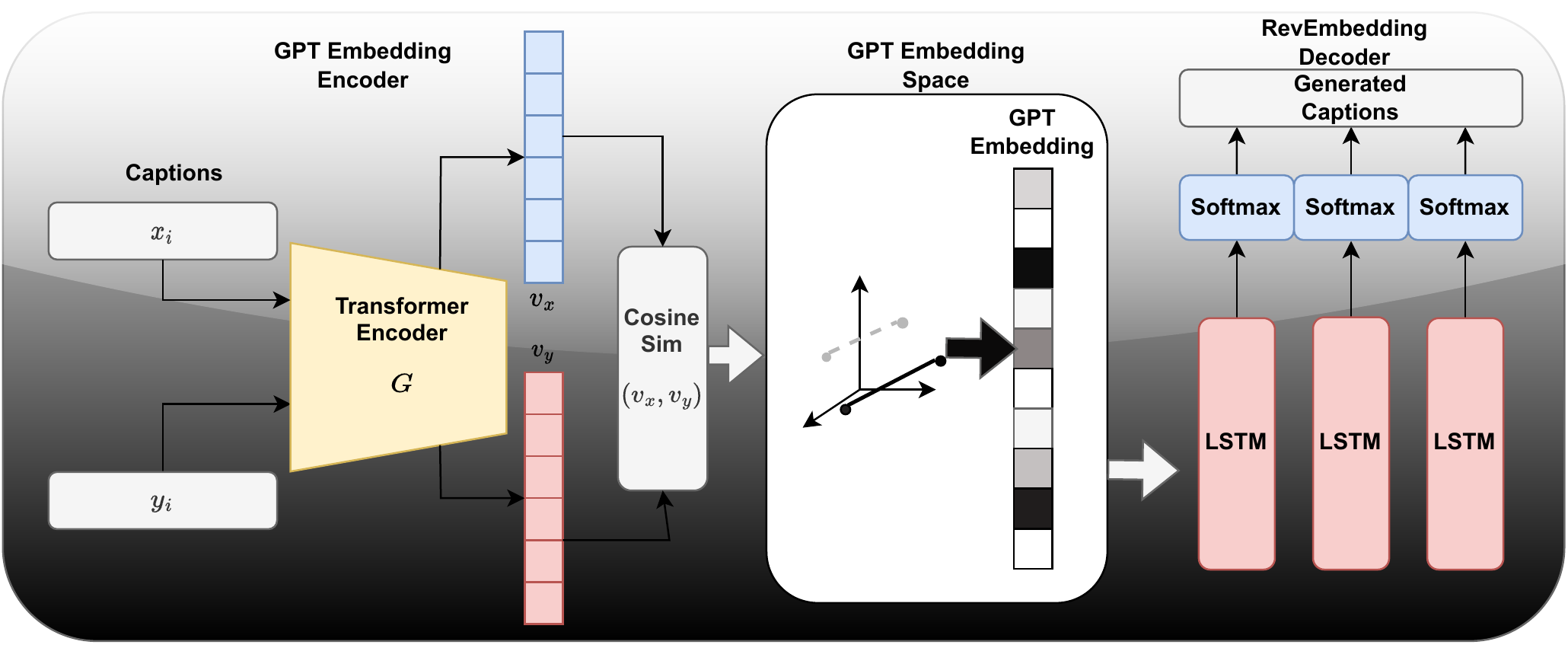}
  \caption{\textbf{Block Diagram demonstrating the embedding mechanism of GPT Embedding and the Reverse Embedding Mechanism of our Proposed Framework.} The GPT Embedding Model uses a Transformer that takes in two texts $x$ and $y$ , for each instance $i$, and calculates the cosine similarity between them. Hence, training such a language model over a large corpus generates the GPT Embedding Space which adequately captures the contextual relationship among texts.}
  \label{fig:Concept}
\end{figure*}

\section{Literature Survey}
\paragraph{Deep Brain Reconstruction:} Deep generative models have been pivotal in advancing visual reconstruction, particularly in deep learning. Prominent models such as Variational Autoencoders (VAEs), Generative Adversarial Networks (GANs), and Latent Diffusion Models (LDMs) have been extensively employed to reconstruct entire images. The standard approach involves using pre-trained deep generative models to learn mappings reconstructing the corresponding latent variables from brain signals. This approach has successfully reconstructed various images, including faces, single-object-centered images, and complex scenes.

Previous studies in visual stimulus reconstruction have primarily focused on the Generic Object Decoding and Deep Image Reconstruction datasets, which are derived from the ImageNet dataset and involve training and testing images with varying fMRI repetitions. Notable research has emerged from this line of investigation. The initial problem was addressed by Shen et al. (2019) through an optimization method utilizing a deep generator network and fMRI-decoded CNN features\cite{Shen2019}. They optimized image pixel values to align with fMRI-decoded features. Beliy et al. (2019) introduced supervised and unsupervised training approaches for fMRI-to-image reconstruction networks to address the scarcity of labeled data, allowing training on "unlabeled" data without fMRI or images\cite{Beliy2019}. Mozafari et al. (2020) recognized the issue of unrecognizable objects in reconstructed images due to an emphasis on pixel-level similarity. They proposed the BigBiGAN-based reconstruction model, which focused on preserving object recognition\cite{Mozafari2020}. Ren et al. (2021) tackled limitations in fMRI data, such as low signal-to-noise ratio and limited spatial resolution, with a Dual VAE-GAN model that learned visually guided latent cognitive representations from fMRI signals and reconstructed image stimuli\cite{Ren2021}. Ozcelik et al. (2022) addressed the challenge of simultaneously reconstructing low-level and high-level image features. They introduced the Instance-Conditioned GAN model, which captured precise semantics and poses information\cite{Ozcelik2022}. Chen et al. (2022) addressed the lack of fMRI-image pairs and effective biological guidance, leading to blurry and semantically meaningless reconstructions\cite{Chen2022}. Their solution involved a sparse masked brain modeling approach and a double-conditioned diffusion model for establishing a precise and generalizable connection between brain activity and visual stimuli.

In recent years, Allen et al. (2022) introduced the Natural Scenes Dataset (NSD) \cite{Allen2022} as a benchmark for fMRI-based natural scene reconstruction. Lin et al. (2022) adapted the StyleGAN2 model using the Lafite framework for text-to-image generation\cite{Lin2022}, while Takagi et al. (2022) and Gu et al. (2022) utilized Stable Diffusion and an improved IC-GAN framework for image reconstruction and pose estimation, respectively\cite{Takagi2022, Gu2022}.

\section{Conceptual Background}
\paragraph{Word Embedding:} Word embedding is a fundamental technique employed in natural language processing (NLP) to represent words in a continuous and low-dimensional vector space(Please refer to \cref{fig:Concept}), referred to as the latent space. Its purpose is to capture the semantic and syntactic relationships among words based on their contextual usage within a vast text corpus.

In contrast to the traditional approach of representing words in NLP using one-hot encoding, where words are represented as sparse binary vectors, word embedding overcomes this limitation by mapping words into a dense vector space. This transformation allows for the capture of meaningful relationships between words, as the spatial proximity of word vectors reflects the semantic similarity between the corresponding words. This proximity arises from the observation that words appearing in similar contexts tend to have similar vector representations.

Popular algorithms for generating word embeddings include Word2Vec \cite{Mikolov2013}, GloVe \cite{Pennington2014}, and FastText \cite{Bojanowski2016}, which leverage large text datasets to learn word representations. GPT, a prominent language model developed by OpenAI \cite{Neelakantan2022}, utilizes word embeddings as an integral component of its architecture. GPT employs contextual word embeddings, also known as contextualized representations, which capture the context-dependent meaning of words. Unlike traditional word embeddings, which assign fixed vector representations to each word, contextual word embeddings in GPT consider the target word, its neighboring words, and the overall sentence or document to generate word representations. The word embedding mechanism in GPT is built upon the Transformer architecture, a deep neural network model designed explicitly for sequence-to-sequence tasks. GPT employs a multi-layer Transformer encoder to process input text and generate contextualized word representations. 

\paragraph{Reverse Embedding:} Reverse word embedding, also called word decoding or word reconstruction, is a crucial process in converting a numerical representation, typically in the form of a vector, back into its corresponding word or textual representation(Please refer to \cref{fig:Concept}). The inverse operation of word embedding maps continuous vector representations within a high-dimensional space back onto its respective word or text \cite{Zoph2016}.

Reverse word embedding plays a significant role in various applications, including language generation, machine translation, and text summarization, where the generated or translated text needs to be transformed back into its original word form. By reconstructing words from their numerical representations, reverse word embedding bridges the continuous vector space and the discrete word space \cite{Vijay2014}.

The underlying principle of reverse word embedding lies in comprehending how word embeddings are learned and utilized. It relies on the associations and relationships learned within the embedding space to reconstruct the original words. This inversion process entails finding the word closest to a given vector representation in the embedding space. Different approaches can be employed, such as nearest neighbor search or computing the cosine similarity between the vector and all the word embeddings within a pre-trained embedding model.

\begin{figure}[ht]
  \centering
  \includegraphics[width=0.8\linewidth]{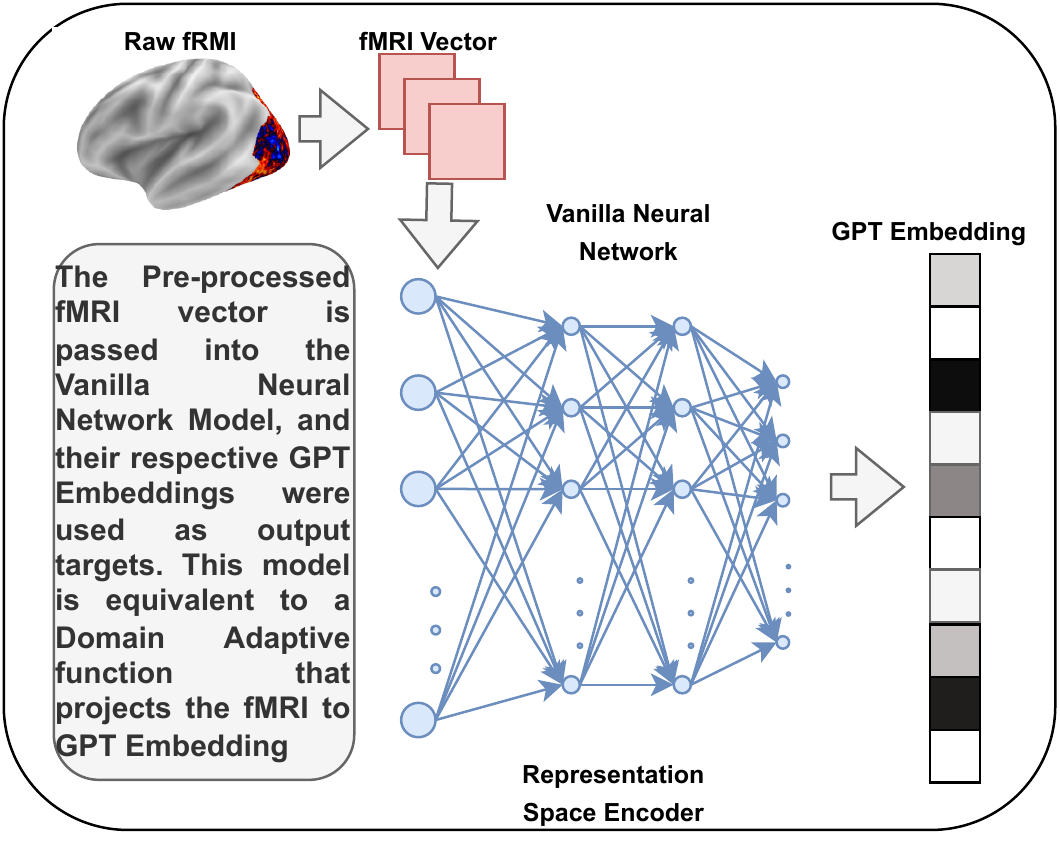}
  \caption{\textbf{Illustrative example for the RSE model.} The RSE model uses a vanilla Neural Network that uses a Mean Squared Error Loss(MSE) to learn the GPT Embedding Space from the input fMRI space.}
  \label{fig:Encoder}
\end{figure}

\section{Proposed Framework}
The details graphical illustration of the DreamCatcher Framework can be found in \cref{fig:How}.
\subsection{Representation Space Encoder}
The first module of the DreamCatcher framework is the Representation Space Encoder(Please refer to \cref{fig:Encoder}), which facilitates the conversion of preprocessed fMRI vectors into the GPT Embedding Space (GPTES). To obtain the fMRI vectors, the Natural Scenes Dataset \cite{Allen2022} is utilized. In contrast, the corresponding captions are obtained from the MS COCO dataset \cite{Lin2014}. These captions undergo conversion into GPT Embeddings via the Embedding API provided by OpenAI.

In contrast to traditional word embeddings, the GPTES incorporates the contextual meaning of words. It is trained using a contrastive objective on paired data. Using cosine similarity, Transformer Encoder $G(.)$ calculates the appropriate distance between a given training pair ($x_i$, $y_i$). This similarity measure ensures the preservation of contextual meaning within the GPTES.

\begin{equation}
    v_{x} = G(x_i)
\end{equation}
\begin{equation}
    v_{y} = G(y_i)
\end{equation}

The Transformer encoder $G(.)$ maps the given inputs, denoted as $x$ and $y$, to their corresponding embeddings, namely $v_x$ and $v_y$, respectively. The similarity between two inputs is quantitatively assessed by measuring the cosine similarity between their respective embeddings, $v_x$ and $v_y$. The cosine similarity metric provides a measure of the directional similarity between the two vectors in the embedding space. It determines the cosine of the angle between the vectors, which ranges from $-1$ (indicating complete dissimilarity) to $1$ (representing perfect similarity).

\begin{equation}
    sim(x,y) = \frac{v_x.v_y}{||v_x||.||v_y||}
\end{equation}

The notion of similarity in terms of word meanings and contextual cues is a universally observed phenomenon. It applies across various domains, encompassing different languages and image-based contextual similarities. Based on this assumption, we propose the utilization of GPTES as a potential candidate for the Brain Representation Space through the concept of Domain Adaptability. It is important to note that the primary objective of the DreamCatcher framework is to generate meaningful captions from fMRI data. The framework does not focus on developing a biologically plausible representation space or a biologically interpretable model.

\begin{figure*}[ht!]
     \centering
     \begin{subfigure}[b]{0.49\textwidth}
         \centering
         \includegraphics[width=\textwidth]{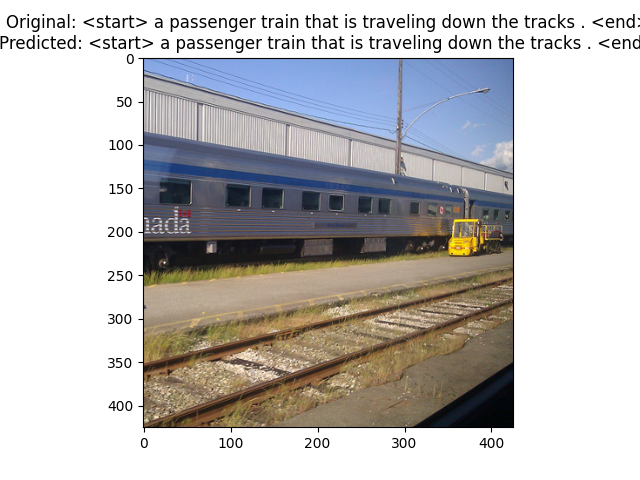}
         \caption{}
         \label{fig:res1}
     \end{subfigure}
     \hfill
     \begin{subfigure}[b]{0.49\textwidth}
         \centering
         \includegraphics[width=\textwidth]{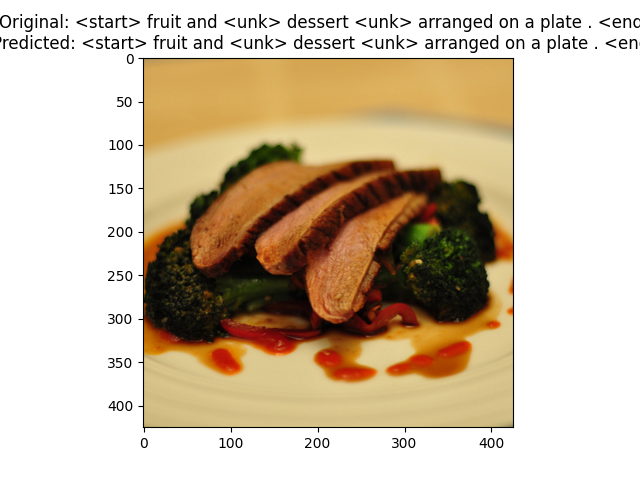}
         \caption{}
         \label{fig:res2}
     \end{subfigure}
     \hfill
     \begin{subfigure}[b]{0.49\textwidth}
         \centering
         \includegraphics[width=\textwidth]{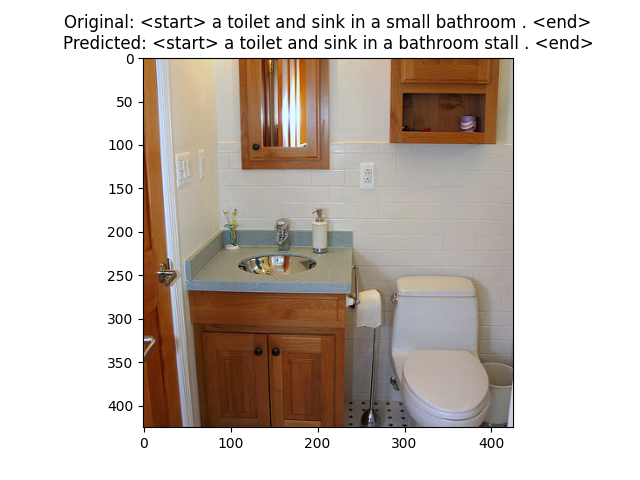}
         \caption{}
         \label{fig:res3}
     \end{subfigure}
     \hfill
     \begin{subfigure}[b]{0.49\textwidth}
         \centering
         \includegraphics[width=\textwidth]{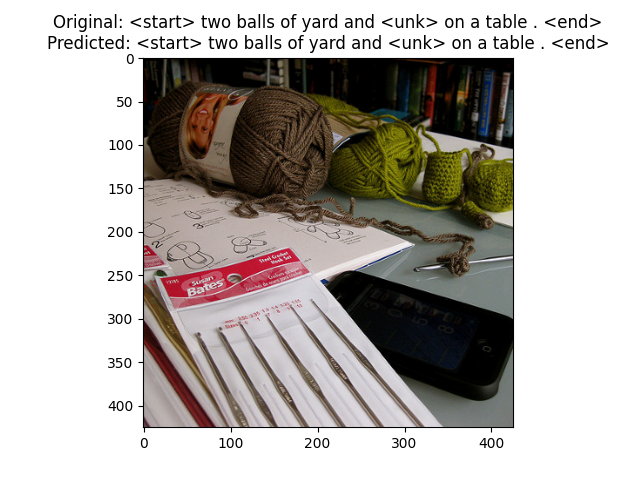}
         \caption{}
         \label{fig:res4}
     \end{subfigure}
     \hfill
        \caption{\textbf{Illustrative Example of Original and Predicted Caption for each Visual Stimuli.} The similarity in text can be observed directly by comparing the captions, which clearly proves the feasibility of fMRI Captioning.}
        \label{fig:result}
\end{figure*}

\subsection{RevEmbedding Decoder}
The second component of the DreamCatcher Framework is the RevEmbedding Decoder, which is responsible for converting GPT Embeddings into captions. However, due to the absence of a Reverse Embedding API in OpenAI, approximations such as Nearest Neighbour are utilized, significantly compromising the accuracy of the generated captions. Hence, the RevEmbedding Decoder model assumes paramount importance in generating captions from the embeddings predicted by the Representation Space Encoder.

In the initial stage, the embeddings-caption pairs generated in the preceding step undergo preprocessing and are stored to form a custom dataset. Subsequently, a Vocabulary module is defined to facilitate the creation and mapping of words to indices. This class also encompasses a function that constructs the vocabulary based on the captions present in the dataset. The function tokenizes the captions, calculates token frequencies, and filters out infrequent words. Additionally, special tokens such as `$<$pad$>$,' `$<$start$>$,' `$<$end$>$,' and `$<$unk$>$' are incorporated into the vocabulary.

Finally, a sequential one-to-many Long Short Term Memory(LSTM) module is trained on the embedding-vocabulary pair to generate the target captions. . It receives GPT Embeddings and caption sequences as input and trains as a one-to-many model. The resulting output corresponds to the predicted captions based on the fMRI input to the Representation Space Encoder.

\begin{figure*}[ht]
  \centering
  \includegraphics[width=0.8\textwidth]{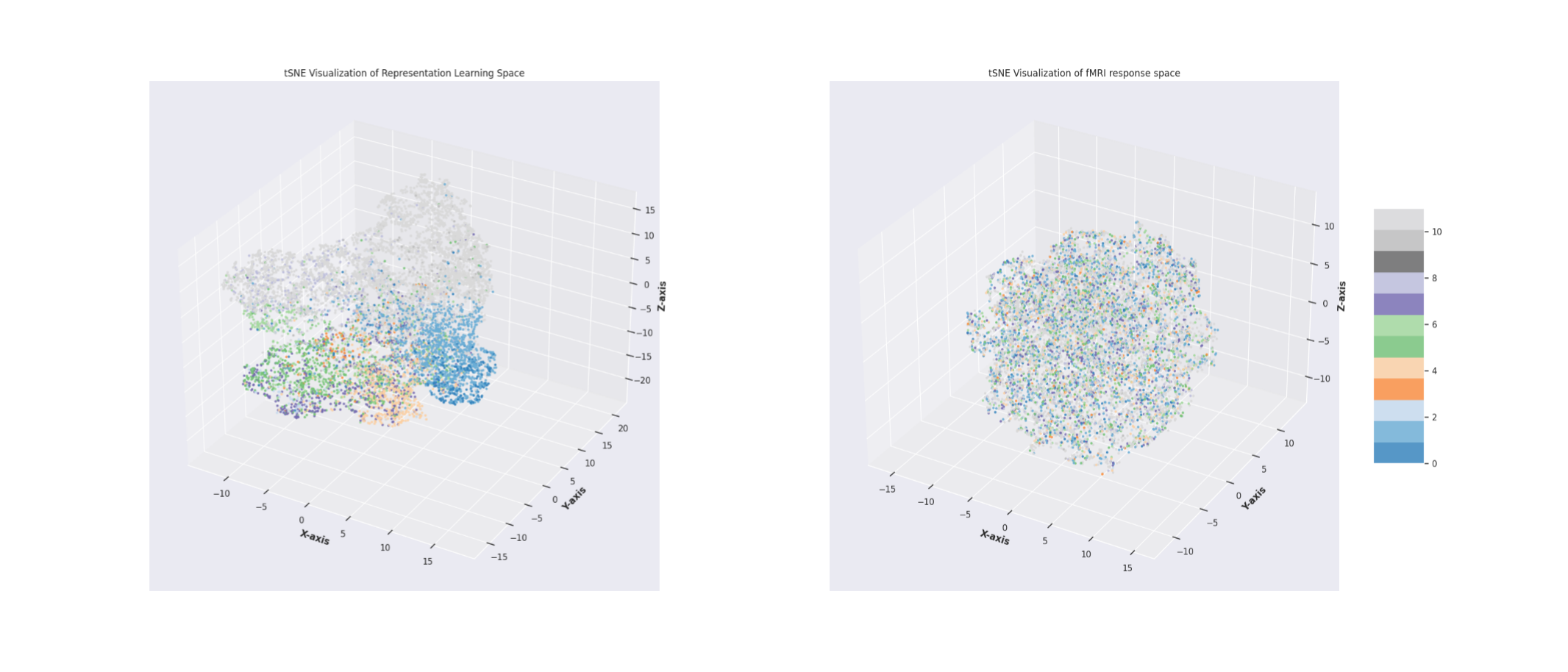}
  \caption{\textbf{t-SNE plot for GPT Embedding Space(Left) and input fMRI Space(Right).} It can be visually observed that GPT Embedding Space provides a category-wise segregation and therefore is a better latent representation in comparison to the input fMRI Space. The segregation also demonstrates that the Embedding Space adequately captures the relationship among the fMRI class labels. }
  \label{fig:tsne}
\end{figure*}

\begin{table*}[]
\begin{tabular}{llccclccc}
\hline
                &               & \multicolumn{3}{c}{Subject 1}                                                              &  & \multicolumn{3}{c}{Subject 2}                                                              \\ \cline{3-5} \cline{7-9} 
Encoder-Decoder & GPT Embedding & \multicolumn{1}{l}{Sentence$\uparrow$} & \multicolumn{1}{l}{Meteor$\uparrow$} & \multicolumn{1}{l}{Perplexity$\downarrow$} &  & \multicolumn{1}{l}{Sentence$\uparrow$} & \multicolumn{1}{l}{Meteor$\uparrow$} & \multicolumn{1}{l}{Perplexity$\downarrow$} \\ \hline
-               & -             & 0.253                        & 0.179                      & 3.486                          &  & 0.241                        & 0.157                      & 3.845                          \\
\checkmark      & -             & 0.276                        & 0.183                      & 1.803                          &  & 0.247                        & 0.167                      & 1.952                          \\
\textbf{\checkmark}       &   \textbf{\checkmark}            & \textbf{0.451}               & \textbf{0.323}             & \textbf{1.024}                 &  & \textbf{0.422}               & \textbf{0.308}             & \textbf{1.037}                 \\ \hline
\end{tabular}
\caption{\textbf{Detailed Results of Component Analysis of each Component of DreamCatcher Framework.} The $\uparrow$ and $\downarrow$ represents the desired value for each metric.}
\label{tab:result}
\end{table*}

\section{Experimental Result and Analysis}
\subsection{Dataset}
\paragraph{Natural Scene Dataset:} A comprehensive account of the Natural Scenes Dataset (NSD), specifications, and data acquisition procedures can be found in a publication by Allen et al. in Nature Neuroscience (2021) \cite{Allen2022}. The NSD dataset encompasses functional Magnetic Resonance Imaging (fMRI) measurements collected from 8 participants who were presented with a substantial number of distinct color natural scenes, ranging from 9,000 to 10,000 images (22,000 to 30,000 trials) across 30 to 40 scan sessions. The fMRI scanning was conducted at 7T using whole-brain gradient-echo Echo Planar Imaging (EPI) at a resolution of 1.8 mm and a repetition time of 1.6 seconds.

The images utilized in the NSD dataset were sourced from the Microsoft Common Objects in Context (COCO) database, square-cropped, and displayed at a size of 8.4° x 8.4°. Out of the total images, a specific set of 1,000 images was shared across all participants, while the remaining images were mutually exclusive for each participant. The images were presented for a duration of 3 seconds, with 1-second gaps between successive images. During the scanning sessions, the participants fixated centrally and engaged in a continuous long-term recognition task related to the presented images.

Pre-processing of the fMRI data involved performing temporal interpolation to correct slice time differences and spatial interpolation to account for head motion artifacts. Subsequently, a general linear model was employed to estimate single-trial beta weights. The NSD dataset also encompasses cortical surface reconstructions generated using FreeSurfer, with both volume- and surface-based versions of the beta weights being created for further analysis and interpretation.

\paragraph{MS COCO Dataset:} The MS COCO (Microsoft Common Objects in Context) \cite{Lin2014} dataset has emerged as a significant benchmark for advancing the field of computer vision. It provides a diverse and comprehensive collection of annotated images, enabling the development and evaluation of a wide range of vision tasks, including object detection, segmentation, and image captioning. MS COCO Captions provide detailed textual descriptions for the images in the MS COCO dataset. With over 330,000 images, each accompanied by multiple human-generated captions, this annotation aspect of MS COCO has revolutionized image captioning research. The captions capture the salient objects, their attributes, and contextual information concisely and descriptively. They serve as a valuable resource for training and evaluating image captioning models, pushing the boundaries of image understanding and natural language processing. By bridging the gap between visual and textual domains, MS COCO captions have enabled advancements in generating human-like image descriptions, opening up new avenues for multimodal research.

\subsection{Feasiblity test of fMRI Captioning}

The potential of generating textual descriptions directly from fMRI data offers a promising avenue for understanding brain activity and decoding mental representations. A feasibility test was conducted using a dataset of fMRI recordings obtained from 8 participants engaged in visual stimulus tasks(Natural Scenes Dataset). The fMRI data were preprocessed and subjected to feature extraction techniques to capture relevant brain activity patterns. Preliminary results from the feasibility test showed promising performance of the fMRI captioning model. The generated captions exhibited reasonable coherence and semantic relevance, capturing essential aspects of the visual stimuli. The graphical illustration can be found in \cref{fig:result}.

\subsection{Efficacy of DreamCatcher Framework}
Since fMRI captioning is still a nascent field, directly comparing our framework with state-of-the-art models was not feasible. Nevertheless, we conducted a comprehensive evaluation to assess the effectiveness of our framework. The evaluation of our proposed framework involved testing our model on two subjects and employing several metrics, including METEOR, Sentence, and Perplexity. The results of our evaluation demonstrated promising performance and the potential of our framework in generating accurate and coherent captions from fMRI data. The METEOR metric, which measures the quality of generated captions by comparing them to reference captions, indicated favorable scores across all subjects. The Sentence metric, which evaluates the syntactic and semantic similarity between generated and reference captions, also showed encouraging results. The detailed result for our model can be obtained in \cref{tab:result}.

\subsection{Verification of GPT Embedding Space Applicability}
To evaluate the ability of our framework to capture adequate representations, we conducted a visualization of the generated representation space using PCA and t-SNE. This analysis aimed to provide insights into the distribution and clustering of the representations, indicating the framework's capacity to capture and differentiate between various features and attributes effectively.

The visualization of the generated representation space revealed encouraging results. The representations exhibited clear separation and distinct clusters, suggesting that our framework successfully captured the relevant information from the input fMRI data. The graphical illustration can be found in \cref{fig:tsne}.

\section{Conclusion}

The human brain's exceptional proficiency in visual processing, including image recognition and scene summarization, have been the subject of extensive research. However, despite significant efforts, a comprehensive understanding of the fundamental mechanisms underlying human visual processing still needs to be discovered. Functional magnetic resonance imaging (fMRI) has emerged as a valuable tool for investigating the neural activity of the human brain, leading to advancements in brain decoding techniques from primary classification approaches to more sophisticated fMRI-to-Image reconstruction methods. 
The contributions of this research include the introduction of fMRI captioning as an alternative approach to fMRI-to-Image Reconstruction, the proposal and feasibility testing of the DreamCatcher framework, and the validation of the GPT Embedding space as a suitable brain representation space for fMRI captioning tasks.

In summary, the advancements in fMRI-based brain decoding techniques, particularly in fMRI captioning, provide valuable insights into the neural mechanisms underlying human visual processing. The DreamCatcher framework represents a significant step forward in capturing the rich cognitive processes involved in visual perception, and its versatility makes it adaptable to other modalities. Continued research in this area holds great promise for expanding our understanding of the human brain and improving various applications, ranging from cognitive neuroscience to educational practices and human-computer interaction. 

{\small
\bibliographystyle{ieee_fullname}
\bibliography{egbib}
}

\end{document}